# Development of a Collaborative Robotic Arm-based Bimanual Haptic Display

Joong-Ku Lee[1], DongHyeon Kim[2], Seong-Su Park', Jiye Lee[2], and Jee-Hwan Ryu[1]

[1] *Department of Civil and Environmental Engineering, KAIST, Daejeon, South Korea*

[2] *Robotics Program, KAIST, Daejeon, South Korea*

(Email: jhryu@kaist.ac.kr)

**Abstract** --- **This paper presents a bimanual haptic display based on collaborative robot arms. We address the limitations of existing robot arm-based haptic displays by optimizing the setup configuration and implementing inertia/friction compensation techniques. The optimized setup configuration maximizes workspace coverage, dexterity, and haptic feedback capability while ensuring collision safety. Inertia/friction compensation significantly improve transparency and reduce user fatigue, leading to a more seamless and transparent interaction. The effectiveness of our system is demonstrated in various applications, including bimanual bilateral teleoperation in both real and simulated environments. This research contributes to the advancement of haptic technology by presenting a practical and effective solution for creating high-performance bimanual haptic displays using collaborative robot arms.**

**Keywords: Bimanual haptic display, Collaborative robot arm, Human-robot interaction**

## 1 INTRODUCTION

The growing interest in the Metaverse and Extended Reality (XR), coupled with ambitious projects like ALOHA [1] that aim to replicate human bimanual capabilities, has highlighted the increasing need for advanced bimanual haptic displays. These devices, which track human movements and deliver realistic force feedback, are crucial for creating immersive and interactive experiences.

However, current commercially available bimanual haptic displays face several challenges. They often struggle to cover the entire human arm workspace and provide adequate force/torque feedback. Configuring a bimanual system using two existing devices is feasible, but these setups are often confined to desktop environments and provide limited feedback, making them unsuitable for complex bimanual tasks. Devices capable of encompassing the full human arm workspace often require substantial dedicated space, limiting their practicality. Exoskeleton-based solutions, while promising, remain largely confined to research settings due to their limited commercial availability and development complexities [2-3].

Previous research has explored using collaborative robot arms as a foundation for bimanual haptic displays. Notable examples include the German Aerospace Center's (DLR) system utilizing DLR/KUKA LWR arms [4-5] and the ANA Avatar XPRIZE-winning NimbRo team's implementation with Franka Emika Panda robots [6]. However, these prior works often overlooked crucial aspects such as optimizing the robot setup configuration to balance workspace, force feedback capabilities, and safety considerations for human-robot interaction. Additionally, they did not adequately address the inherent inertia and friction of the robot arms, which can significantly reduce the realistic and transparent haptic experience.

This paper aims to bridge these gaps by presenting a comprehensive approach to developing a collaborative robot arm-based bimanual haptic display. We detail the implementation of essential techniques, including setup configuration optimization and friction/inertia compensation, to transform collaborative robots into effective haptic devices. We also provide a demo video showcasing the system's capabilities and potential applications across various domains.

## 2 SYSTEM DESIGN AND IMPLEMENTATION

The proposed bimanual haptic display utilizes two

Franka Emika Panda robot arms. This section details the core technologies that enable the use of collaborative robot arms as haptic displays: setup configuration optimization and inertia/friction compensation.

### 2.1 Setup configuration optimization

The performance of a collaborative robot-based bimanual haptic display is significantly influenced by its setup configuration, which encompasses the base position and orientation of the robot arms, as well as the grab angle between the human hand and the robot end-effector.

The setup configuration impacts factors such as workspace and haptic feedback force/torque, which directly affect the overall performance of the haptic display. To address this, [7] proposed an optimization scheme to identify the best-performing setup configuration. The optimization process considers several key factors:
- Workspace coverage: The haptic display should cover the human arm workspace as comprehensively as possible to facilitate seamless and natural human arm movements.
- Redundancy: Robot arm redundancy is crucial for enabling dexterous human arm movements and mitigating the risk of human-robot collisions.
- Renderable haptic feedback force/torque: The robot arms must be capable of delivering sufficient haptic feedback force/torque across all directions to ensure realistic and immersive interaction.
- Human-robot collision: The optimization process prioritizes minimizing the potential for collisions between the human operator and the robot arms, ensuring a safe user experience.

The optimized setup configuration for the Franka Emika Panda robot arm is presented in Table 1. The practical implementation of this setup configuration is visually depicted in Fig. I. Our bimanual haptic display is built upon this setup configuration.

Table I Optimized design variables with the Franka Emika Panda robot arms

| Design Variables | Values | Units |
|---|---|---|
| Base Position ($X_b$) | (-0.205, 0.066, 0.262) | m |
| Base Orientation ($R_b$) | (-0.900, 0.177, -0.219) | rad |
| Grab Angle ($\theta_{grab}$) | -0.569 | rad |

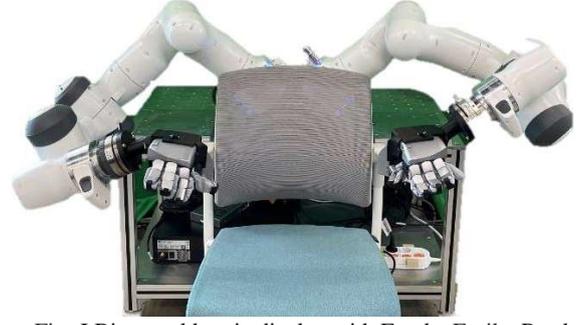

Fig. I Bimanual haptic display with Franka Emika Panda robot arm with optimized setup configuration

### 2.2 Inertia/friction Compensation

One of the primary challenges in employing industrial collaborative robots as haptic displays lies in *their* relatively high inertia and friction levels compared to dedicated haptic devices. The presence of inertia and friction can adversely affect the accuracy and precision of haptic feedback, potentially leading to increased user fatigue, particularly during extended periods of operation.

To address the inertia challenge, we implement the inertia reshaping method proposed in [8]. The fundamental principle of this method is to employ joint torque feedback as a control input to modify the effective inertia of the robot arm. The dynamics of a flexible joint robot (FJR) can be represented as:

$$B\ddot{\theta} + \tau_j = \tau_m + \tau_f \quad (1)$$
$$M(q)\ddot{q} + C(q,\dot{q})\dot{q} + g(q) = \tau_j + \tau_{ext}, \quad (2)$$

where $B$ denotes the inertia of the motor, $\tau_j$ is the torque applied to the joint, $\tau_m$ is the control input, $\tau_f$ is the frictional torque, and $\tau_{ext}$ is the external torque applied to the joint. By applying the control input $\tau_m$ as:

$$\tau_m = \tau_j + BB_d^{-1}(-\tau_j + u), \quad (3)$$

where $B_d$ is the desired motor inertia, the FJR robot dynamics are transformed into:

$$B_d\ddot{\theta} + \tau_j = u + B_d B^{-1} \tau_f \quad (4)$$
$$M(q)\ddot{q} + C(q,\dot{q})\dot{q} + g(q) = \tau_j + \tau_{ext}, \quad (5)$$

The resulting equation demonstrate the reshaped inertia of the system and the modified friction level at the robot joint. The selection of the reshaping ratio $BB_d^{-1}$ is a key design parameter. In our system, we empirically determined this ratio to be 3 for translational movement and 4 for rotational movement to achieve a balance between responsiveness and stability.

While the inertia reshaping method effectively addresses the inertia challenge, residual friction within the system can still impact the performance of the haptic display. To mitigate this, we employ an energy-based friction compensation method proposed in [9]. This method leverages the relationship between input energy, stored energy, and energy dissipated through friction to adaptively estimate and compensate for the time-varying friction coefficient. The core principle behind this approach is the observation that when the stored energy in the system is zero, there exists a clear relationship between the net energy input and the energy dissipated by friction, allowing for friction estimation without requiring explicit model information.

This energy-based friction compensation method offers two key advantages when applied to collaborative robot arms:

1. Independence from robot system model information: The method estimates friction based on the energy levels observed during human-robot interaction, eliminating the need for precise robot model identification.
2. Ease of online implementation: It directly estimates friction coefficients for a given friction model without requiring any system-specific tuning parameters, simplifying the implementation process.

We apply this energy-based friction compensation method on top of the inertia reshaping method. The result from single degree-of-freedom (DOF) experiment demonstrate a significant reduction in the interaction torque required to move a single robot joint when both methods are combined (blue line in Fig. 2). Furthermore, the combined approach outperforms each method applied individually (purple, green line in Fig. 2), highlighting the effectiveness of their synergistic integration.

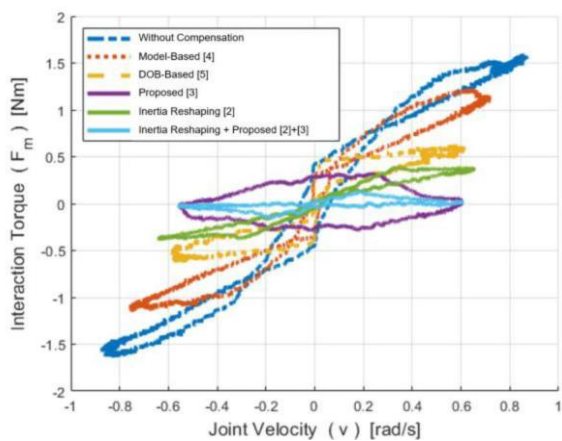

Fig. 2 Joint velocity vs. interaction torque curve with different friction compensation methods. The combination of inertia reshaping and proposed energy-based friction compensation method outperforms other methods.

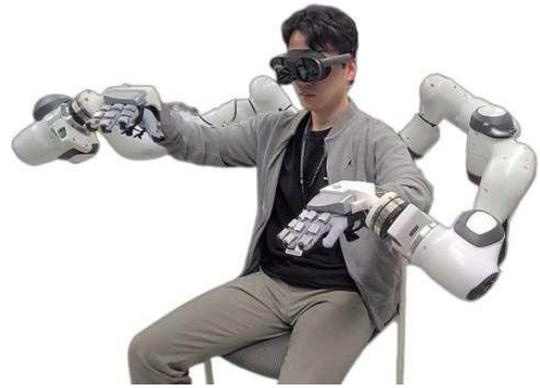

Fig. 3 A proposed bimanual haptic display combined with a haptic glove and a YR headset

## 3 RESULTS

The proposed bimanual haptic display can be seamlessly integrated with a variety of human haptic interfaces, such as haptic gloves and virtual reality (VR) headsets. Fig. 3 illustrates a potential configuration showcasing our proposed bimanual haptic display combined with a haptic glove and a VR headset. Through its dexterity and transparency, our system is expected to provide users with a highly realistic and immersive experience.

As demonstrated in Fig. 4 and Fig. 5, our bimanual haptic display can be effectively applied to diverse scenarios, including bimanual bilateral teleoperation and bilateral teleoperation within a simulated environment. The system's extensive workspace coverage and sophisticated force feedback capabilities enable it to perform a wide range of tasks that humans typically execute in real-life settings.

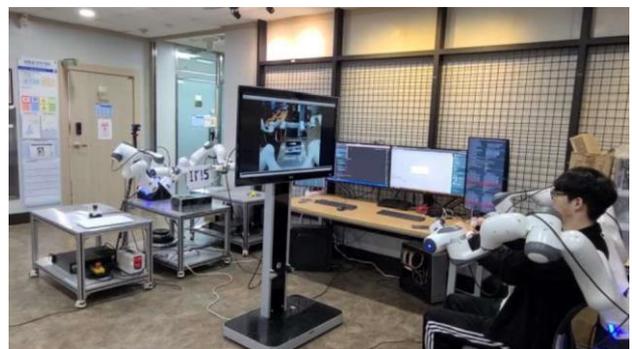

Fig. 4 Application of the proposed bimanual haptic display to bimanual bilateral teleoperation

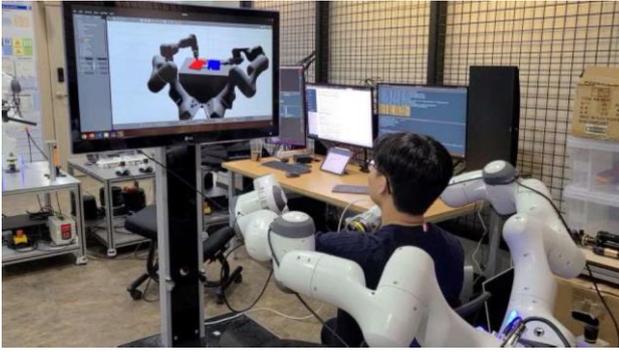

Fig. 5 Application of the proposed bimanual haptic display to bilateral teleoperation in a simulated environment

## 4 CONCLUSION

This paper introduces a collaborative robot arm-based bimanual haptic display that overcomes the limitations of existing devices. By optimizing the setup configuration and implementing inertia/friction compensation, we have significantly enhanced the system's performance and user experience. The proposed system's dexterity, transparency make it suitable for various applications, as shown in results.

Future research will focus on refining the system's capabilities and exploring its integration with diverse haptic interfaces to create even more immersive and interactive experiences.

ACKNOWLEDGEMENT

This work was supported in part by the Robot Industry Core Technology Development Program (20023294, Development of shared autonomy control framework and AI-based application technology for enhancing tasks of hyper realistic telepresence robots in unstructured environment) funded by the Ministry of Trade, Industry & Energy (MOTIE, Korea), and in part by the Field-oriented Technology Development Project for Customs Administration through National Research Foundation of Korea(NRF) funded by the Ministry of Science \& ICT and Korea Customs Service(2022M311Al09507521).